\documentclass[journal]{IEEEtran}
\usepackage{amsmath,amsfonts}
\usepackage{algorithm}
\usepackage{array}
\usepackage{etoolbox}
\usepackage[caption=false,font=footnotesize]{subfig}
\usepackage{textcomp}
\usepackage{stfloats}
\usepackage{url}
\usepackage{verbatim}
\usepackage{graphicx}
\usepackage{cite}
\usepackage{pythonhighlight}
\usepackage[acronym]{glossaries}
\usepackage{float}
\usepackage{adjustbox}
\usepackage{multirow}
\usepackage{booktabs}
\usepackage{tabularx}
\usepackage{siunitx}
\usepackage{algpseudocode}
\usepackage{hyperref}
\usepackage{cleveref}

\makeatletter
\patchcmd{\@makecaption}
  {{\normalfont\footnotesize ##1}\\{\normalfont\footnotesize\scshape ##2}}
  {{\normalfont\footnotesize ##1}\\{\normalfont\footnotesize ##2}}
  {}{}
\makeatother

\usepackage{graphicx}
\usepackage{tikz}
\usetikzlibrary{positioning,fit}

\newacronym[description={Markov Decision Process}]{mdp}{MDP}
  {Markov decision process}
\newacronym[description={Width-Based Search under Partially Observing}]{powbs}{PO-WBS}
  {Width-Based Search under Partially Observing}
\newacronym[description={Iterated Width under partial Observability}]{poiw}{PO-IW}
  {Partially Observing Iterated Width}
\newacronym[description={Planning Domain Definition Language}]{pddl}{PDDL}
  {Planning Domain Definition Language}
\newacronym[description={Probability Density Function}]{pdf}{PDF}
  {probability density function}
\newacronym[description={Random Variable}]{rv}{RV}
  {random variable}
\newacronym{rwth}{RWTH}
  {Rheinisch-Westfälische Technische Hochschule}
\newacronym{strips}{STRIPS}
  {Stanford Research Institute Problem Solver}
\newacronym[description={Partially Observable Markov Decision Process}]{pomdp}{POMDP}
  {Partially Observable Markov decision process}
\newacronym[description={Fully Observable Deterministic}]{fod}{FOD}
  {Fully Observable Deterministic}
\newacronym[description={Partially Observable Deterministic}]{pod}{POD}
  {Partially Observable Deterministic}
\newacronym[description={Markov Chain}]{mc}{MC}
  {Markov Chain}
\newacronym[description={probabilistic Width-Based Search}]{pwbs}{PWBS}
  {probabilistic Width-Based Search}
\newacronym{ucb}{UCB}
  {Upper Confidence Bound}
\newacronym{cot}{COT}
  {Chain of thoughts}
\newacronym{tot}{TOT}
  {Tree of thoughts}  
\newacronym{ucbt}{UCBT}
  {Upper Confidence Bound for Trees}
\newacronym{mcts}{MCTS}
  {Monte Carlo Tree Search}
\newacronym{fm}{FM}
  {Foundation Models}
\newacronym{bw}{BW}
  {Blocks World}
\newacronym{riw}{RIW}
  {Rollout Iterated Width}
\newacronym{miw}{m-IW}
  {m-Iterated Width}
\newacronym{rnd}{RND}
  {Random Network Distillation}
\newacronym{bee}{BEE}
  {Boundary Extension Encoding}
\newacronym{ale}{ALE}
  {Arcade Learning Environment}
\newacronym{tamp}{TAMP}
  {Task and Motion Planning}
\newacronym{gp}{GP}
  {Gaussian Process}
\newacronym{rl}{RL}
  {Reinforcement learning}
\newacronym{ml}{ML}
  {Machine Learning}
\newacronym{mse}{MSE}
  {Mean Squared Error}
\newacronym{llm}{LLM}
  {Large Language Model}
\newacronym{vlm}{VLM}
  {Vision Language Model}
\newacronym{wbp}{WBP}
  {Width-Based Planning}
\newacronym{iw}{IW}
  {Iterated Width}
\newacronym{siw}{SIW}
  {Serialized Iterated Width}
\newacronym{bfs}{BFS}
  {Best First Search}
\newacronym{brfs}{BRFS}
  {Breadth First Search}
\newacronym{dfs}{DFS}
  {Depth First Search}
\newacronym{gbfs}{GBFS}
  {Greedy Best First Search}
\newacronym{vae}{VAE}
  {Variational Autoencoder}
\newacronym{id}{ID}
  {Inverse Dynamic}
\newacronym{ie}{IE}
  {Intrinsic Exploration}
\newacronym{ir}{IR}
  {Intrinsic Rewards}
\newacronym{ff}{FF}
  {Fast-Forward}
\newacronym{bn}{BN}
  {Bayesian Network}
\newacronym{bo}{BO}
  {Bayesian Optimization}
\newacronym{ldm}{LDM}
  {Latent Diffusion Model}
\newacronym{mi}{MI}
  {Mutual Information}
\newacronym{bnn}{BNN}
  {Bayesian Neural Network}
\newacronym{cpt}{CPT}
  {Conditional Probability Tables}
\newacronym{dag}{DAG}
  {Directed Acyclic Graph}
\newacronym{jpd}{JPD}
  {Joint Probability Distribution}
\newacronym{gbfsw}{GBFS-W}
  {Greedy Best First Search with Width}
\newacronym{wbs}{WBS}
  {Width-Based Search}
\newacronym{bfws}{BFWS}
  {Best First Width Search}
\newacronym[description={Deep Q-Network}]{dqn}{DQN}
  {Deep Q-Network}
\newacronym[description={Kullback Leibler Divergence}]{kld}{KLD}
  {Kullback–Leibler divergence}
\newacronym[description={Prediction Gain}]{pg}{PG}
  {Prediction Gain}
\newacronym[description={Information Gain}]{ig}{IG}
  {Information Gain}
\newacronym[description={Proximal Policy Optimization}]{ppo}{PPO}
  {Proximal Policy Optimization}
\newacronym[description={Actor-Critic}]{ac}{AC}
  {Actor-Critic}
\newacronym[description={Temporal Difference Learning}]{tdl}{TDL}
  {Temporal Difference Learning}
\newacronym[description={Artificial Intelligence}]{ai}{AI}
  {Artificial Intelligence}
\newacronym[description={Neural Network}]{nn}{NN}
  {Neural Network}
\newacronym[description={Convolutional Neural Network}]{cnn}{CNN}
  {Convolutional Neural Network}
\newacronym[description={Recurrent Neural Network}]{rnn}{RNN}
  {Recurrent Neural Network}
\newacronym[description={Long Short-Term Memory}]{lstm}{LSTM}
  {Long Short-Term Memory}
\newacronym[description={Generative Adversarial Network}]{gan}{GAN}
  {Generative Adversarial Network}
\newacronym[description={Support Vector Machine}]{svm}{SVM}
  {Support Vector Machine}
\newacronym[description={Principal Component Analysis}]{pca}{PCA}
  {Principal Component Analysis}
\newacronym[description={Computer Vision}]{cv}{CV}
  {Computer Vision}
\newacronym[description={Natural Language Processing}]{nlp}{NLP}
  {Natural Language Processing}
\newacronym[description={State Space Model}]{ssm}{SSM}
  {State Space Model}
\newacronym[description={Structured State Space Model}]{ss4}{SS4}
  {Structured State Space Model}
\newacronym[description={Multi-Joint Dynamics with Contact}]{mujoco}{MuJoCo}
  {Multi-Joint Dynamics with Contact}
\newacronym[description={Model Predictive Control}]{mpc}{MPC}
  {Model Predictive Control}
\newacronym[description={State Of The Art}]{sota}{SOTA}
  {State-of-the-Art}
\newacronym[description={Segment Anything Model}]{sam}{SAM}
  {Segment Anything Model}
\newacronym[description={Vision Transformer}]{vit}{ViT}
  {Vision Transformer}
\newacronym[description={Contrastive Language–Image Pre-training}]{clip}{CLIP}
  {Contrastive Language–Image Pre-training}
\newacronym[description={Large Language Model Meta AI}]{llama}{LLaMA}
  {Large Language Model Meta AI}
\newacronym[description={Large Language and Vision Assistant}]{llava}{LLaVA}
  {Large Language and Vision Assistant}
\newacronym[description={Joint Embedding Predictive Architecture}]{jepa}{JEPA}
  {Joint Embedding Predictive Architecture}

\newacronym[description={Structured State Representation}]{ssr}{SSR}
  {Structured State Representation}

\newacronym[description={State-Actions-Successor}]{sas}{SAS}
{State-Actions-Successor}

\newacronym[description={International Planning Competition}]{ipc}{IPC}
{International Planning Competition}

\newacronym[description={Lightweight Automated Planning ToolKiT}]{laptk}{LAPKT}
{Lightweight Automated Planning ToolKiT}

\newacronym[description={Vision-Language Interpreter for Robot Task Planning}]{vilain}{ViLaIn}
{Vision-Language Interpreter for Robot Task Planning}

\newacronym[description={Problem Description Generation Dataset}]{prodg}{ProDG}
{Problem Description Generation Dataset}

\newacronym[description={Problem Description}]{pd}{PD}
{Problem Description}

\newacronym{pddlgym}{PDDL Gym}{PDDLGym Environment}
\newacronym{grounder_name}{SYMBOLIZER}{Symbolic States from non-symbolic Observations}
\newacronym{complete_system_name}{VLM-Search}{VLM-grounded Width-Based Search}
\newacronym{system_name}{SYMBOLIZER}{SYMBOLIZER}

\newacronym[description={Generative Pre-training Transformer}]{gpt}{GPT}
{Generative pre-trained Transformer}

\newacronym[description={Generative Pre-training Transformer}]{bert}{BERT}
{Bidirectional Encoder Representations from Transformers}

\newacronym[description={Upper Confidence bounds applied to Trees}]{utc}{UTC}
{Upper Confidence bounds applied to Trees}

\newacronym[description={Range between the 25\% quantile and 75\% quantile}]{iqr}{IQR}
{Interquartile Range}

\newacronym[description={Travelling Salesperson PDDL domain}]{tsp}{TSP}
{Travelling Salesperson}

\newacronym[description={The correct, Ground Truth}]{gt}{GT}
{Ground Truth}

\newacronym[description={Sensor Model}]{sensor_model}{SENSOR_MODEL}{Sensor Model}
\newacronym[description={Prediction Model}]{prediction_model}{PREDICTION_MODEL}{Prediction Model}
\newacronym[description={Novelty Table Real}]{closedlist_real}{CLOSED_LIST_REAL}{Novelty Table (Real)}
\newacronym[description={Novelty Real}]{novelty_real}{NOVELTY_REAL}{Novelty (Real)}
\newacronym[description={Novelty Estimation}]{novelty}{NOVELTY}{Novelty}
\newacronym[description={Novelty Table Estimation}]{closedlist}{NT}{Novelty Table}

\newacronym[description={Precision}]{precision}{PREC}{Precision}
\newacronym[description={Recall}]{recall}{REC}{Recall}
\newacronym[description={F1-Score}]{f1_score}{F1}{F1-Score}

\begin{document}

\title{SYMBOLIZER: Symbolic Model-free Task Planning with VLMs}

\author{Sami Azirar$^{1,2,3}$*, Zlatan Ajanovic$^{1}$*, Hermann Blum$^{2,3}$% <-this % stops a space
\thanks{$^1$RWTH Aachen, $^2$Robot Perception and Learning Lab, University of Bonn, $^3$Lamarr Institute for Machine Learning and Artificial Intelligence, * denotes equal contribution.}
\thanks{This research is partially supported by the Lamarr Institute for Machine Learning and Artificial Intelligence.}
}

\maketitle

\begin{abstract}
Traditional Task and Motion Planning (TAMP) systems depend on physics models for motion planning and discrete symbolic models for task planning. Although physics model are often available, symbolic models (consisting of symbolic state interpretation and action models) must be meticulously handcrafted or learned from labeled data. 
This process is both resource-intensive and constrains the solution to the specific domain, limiting scalability and adaptability. 
On the other hand, Visual Language Models (VLMs) show desirable zero-shot visual understanding (due to their extensive training on heterogeneous data), but still achieve limited planning capabilities. Therefore, integrating VLMs with classical planning for long-horizon reasoning in TAMP problems offers high potential. Recent works in this direction still lack generality and depend on hand-crafted, task-specific solutions, e.g. describing all possible objects in advance, or using symbolic action models. 
We propose a framework that generalizes well to unseen problem instances. The method requires only lifted predicates describing relations among objects and uses VLMs to ground them from images to obtain the symbolic state. Planning is performed with domain-independent heuristic search using goal-count and width-based heuristics, without need for action models. Symbolic search over VLM-grounded state-space outperforms direct VLM-based planning and performs on par with approaches that use a VLM-derived heuristic. This shows that domain-independent search can effectively solve problems across domains with large combinatorial state spaces. We extensively evaluate on extensively evaluate our method and achieve state-of-the-art results on the ProDG and ViPlan benchmarks.
\end{abstract}


Robotic manipulation in unstructured environments requires reasoning over both high-level task objectives and low-level geometric constraints. Task and Motion Planning (TAMP) has emerged as one of the most general and expressive paradigms for addressing this challenge, enabling robots to jointly reason about symbolic actions and continuous feasibility \cite{garrett2021integrated}. By integrating discrete task planning with motion-level feasibility checks, TAMP can represent a wide range of complex manipulation problems, from multi-step assembly to rearrangement in cluttered scenes.

\begin{figure}[!t]
    \centering
    \includegraphics[width=\columnwidth]{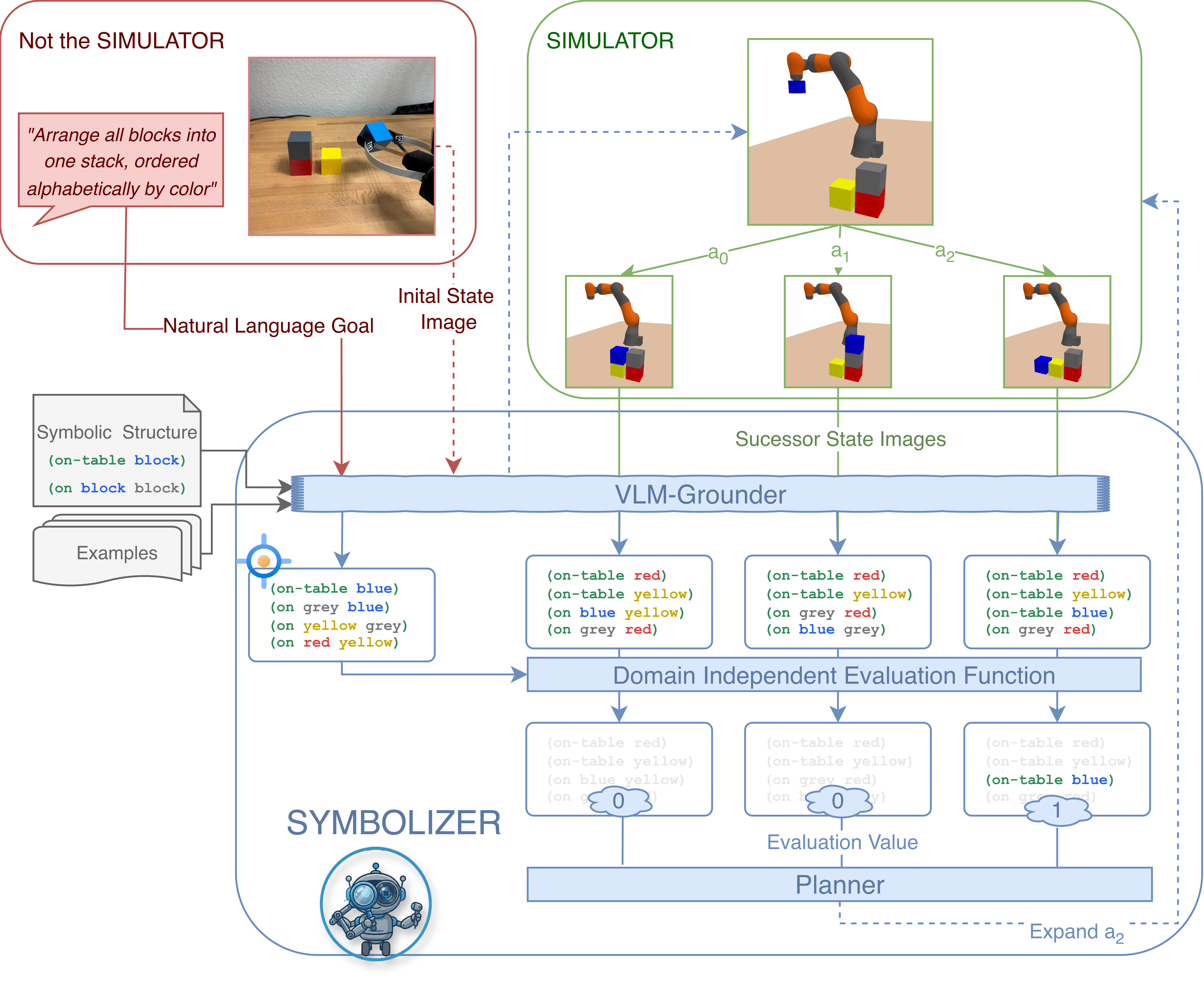}
    \caption{SYMBOLIZER grounds visual observations into well-typed symbolic states using structured VLM outputs, enabling domain-independent planning without handcrafted action models.}
    \label{fig:teaser}
\end{figure}

Despite its generality, classical TAMP approaches face significant limitations. In particular, they rely heavily on manually engineered symbolic representations, including action schemas, predicates, and domain models. Constructing these representations requires substantial domain expertise and often limits scalability to new environments. In parallel, vision-language models (VLMs) have emerged as powerful tools for generating rich semantic representations, enabling robots to interpret natural language instructions and perceptual inputs in open-world settings. Predominantly, current research seeks to exploit these priors to train language-conditioned policies end-to-end \cite{kim2024openvla}, leveraging imitation learning from human demonstrations and reinforcement learning in simulation \cite{luijkx2025llm}. These approaches enable robots to acquire reusable skills that generalize across diverse environments. However, while such methods excel at reactive behaviors and short-horizon tasks, they often struggle with problems that require structured, long-horizon reasoning and explicit planning \cite{li2024behavior1k,kambhampati2024position}.

In this work, we aim to bridge the gap between symbolic planning and learning-based perception by leveraging VLMs for grounding within a TAMP framework. Specifically, we utilize the semantic understanding capabilities of VLMs to map raw sensory observations and textual descriptions into symbolic representations suitable for planning. A common issue for such mappings are hallucinations and structural inaccuracies that cause errors with the strict symbolic requirements of TAMP. We avoid such issues by proposing a formal representation that is compatible with structured output conditioning of VLMs but agnostic to the scenario and task. This enables us to reduce the need for hand-crafted domain models while retaining the compositional reasoning benefits of TAMP.

We propose a unified framework that integrates VLM-based grounding with simulator-driven task and motion planning. Our approach uses VLM grounding to infer symbolic predicates from visual or textual state descriptions, which are then used by a planner to generate candidate action sequences. These plans are validated and refined through simulator-based search, enabling robust execution in complex environments.

In summary, our contributions are:
\begin{itemize}
\item A robust method for grounding symbolic representations and goals from image or textual description using VLMs.
\item A novel framework that combines VLM-based grounding with TAMP and simulator-based search.
\item Extensive experimental evaluation and ablation studies demonstrating the effectiveness of the proposed approach.
\end{itemize}
Code and an interactive demo are available at \url{https://anonymous.4open.science/r/symbolizer-2ECE/}.

\section{RELATED WORK}

Planning is the process of finding a sequence of actions that transforms an initial state into a desired goal state \cite{haslum2019introduction}, where the symbolic action model defines which preconditions must be satisfied for an action to be executed and the effects of its execution. While action models are part of the standard formulation of classical planning problems, effective planning can be done in a simulator, without any access to the action model, given that the symbolic representation of the state is given \cite{francesPurelyDeclarativeAction2017}. Using the blind Width Based Search (WBS) based algorithms, the authors achieved SOTA results in playing Atari using the RAM \cite{WBS-Atari} state or directly pixel based features \cite{WBS-pixels}. To archive more meaningful representations, learning meaningful features is done in \cite{asai2017classical,WBS-learn} which is then used to do classic planning.

Large Language Models (LLM)  demonstrated desired capabilities for open-world task planning (when combined with skill affordances) \cite{brohan2023can}, which can be further integrated into classical planning frameworks \cite{hazra2024saycanpay,lin2023text2motion}, RL guidance on task level \cite{luijkx2025llm}, generating symbolic domain files \cite{smirnovGeneratingConsistentPDDL2024b} for planning or generate directly Behavior Trees \cite{zhou2024llm}, and even partial classical planning capabilities when fine-tuned on planning problems \cite{pallaganiPlansformerGeneratingSymbolic2022}.
However, these approaches remain constrained by the current limitations of LLMs.
and even the most advanced models fail to demonstrate robust planning performance\cite{valmeekamLLMsStillCant2024}, struggle to reason accurately about real-world relations and over long horizons \cite{wang2024can,vafaEvaluatingWorldModel2024}, and, since they are not yet fully interpretable, pose a risk of unpredictable errors when relied upon solely \cite{guan2023leveraging}.

Instead of relying on the LLM's capabilities for solving problems, other approaches use them to bridge the gap between sub-symbolic and symbolic representations. PDDL problem definitions have been created from text \cite{liuLLM+PEmpoweringLarge2023} or images \cite{dangPlanningVisionLanguageModels2025,shirai2024vision} using VLMs. \textsc{ViLaIn} \cite{shirai2024vision}, for instance, chains an open-vocabulary detector with domain-specific object descriptions, a captioning model, and an LLM prompted with the full symbolic action model and few-shot examples to generate PDDL problems, with optional planner-feedback retries to correct errors. Such approaches rely on domain knowledge and manual curation at every stage of the pipeline, limiting their generalizability to new domains. To reduce this engineering effort, \cite{zhangDKPROMPTDomainKnowledge2024} propose DK-PROMPT, which leverages domain knowledge from the PDDL specification to automatically generate queries over predicates and action effects for a VLM. While this removes the need for explicit problem descriptions, it still relies on manually designed query templates and remains limited to simple, type-level predicates (e.g.\ ``is a block on a table''), failing to capture instance-level relations (e.g.\ ``is the red block on the blue block'') and incurring significant query overhead at each planning step. More recently, \textsc{ViPlan} \cite{merler2026viplanbenchmarkvisualplanning} formalizes two complementary paradigms for visual planning with VLMs. In the VLM-as-planner setting, the model directly generates actions from visual observations, conditioned on the current state, goal, and available actions, and is iteratively queried in a closed loop after each execution step. In contrast, the VLM-as-grounder setting integrates the model with a symbolic planner by using it to infer the truth values of grounded predicates from images, both to initialize the state and to validate action effects online while executing symbolic plans.

While other approaches use free-form text, a key requirement for using VLMs as sensor models is the ability to obtain reliable, well-formed structured outputs rather than free-form text. Constrained decoding achieves this by masking invalid tokens at each auto-regressive generation step, so that only sequences conforming to a given formal grammar can be produced. The underlying mechanism has been formulated as finite-state machine transitions over the vocabulary \cite{willard2023efficientguidedgenerationlarge}. Structured outputs are now a standard API option offered by proprietary \cite{google2024gemini_controlled_generation, mistral2024jsonmode} and open-source frameworks \cite{kwon2023efficient}. Although strict grammar constraints have been shown to degrade reasoning by collapsing the model's computation to TC0 \cite{tam-pmlr-v267-banerjee25}, this concern is orthogonal to our use case, which targets perception rather than multi-step inference.

\paragraph*{Research Gap}
\label{sec:research-gap}
The primary research gap addressed in this work is to enable generalized open-world planning using only a physics simulator, without relying on a symbolic action model \cite{bonet2020learning}. Existing approaches that enable classical planning either require the full action model, providing only the initial problem description \cite{liuLLM+PEmpoweringLarge2023,shirai2024vision}, or are limited to simple domains where the VLM can be queried on a type level \cite{zhangDKPROMPTDomainKnowledge2024}. No prior work has used a VLM as a general-purpose sensor model to enable classical planning in complex domains. We hypothesize that, given a lifted predicate vocabulary and a small number of in-context examples, VLMs can ground symbolic states accurately enough for systematic search to find plans, without any domain-specific engineering. To our knowledge, no such approach exists in the literature.

\section{Problem Setting}

\begin{figure}[!t]
    \centering
    \includegraphics[width=\columnwidth]{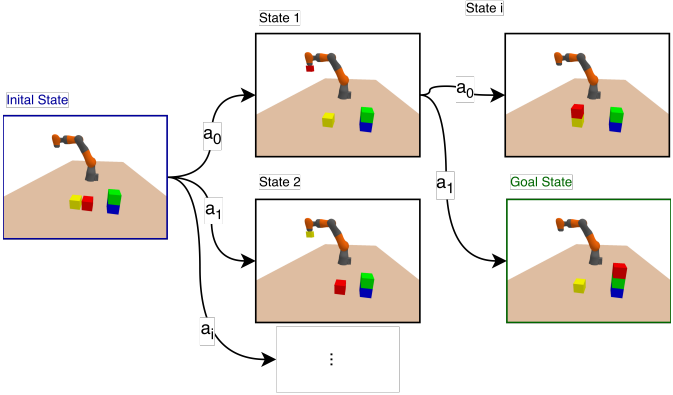}
    \caption{Example of planning with a simulator. From an initial state, the planner expands successor states generated by the simulator, without requiring explicit knowledge of the underlying actions.}
    \label{fig:rollouts}
\end{figure}

We consider Task and Motion Planning under Partial Observability, in domains
where no symbolic action model is provided a priori and the agent receives only
partial, noisy observations of the world.

Let $x \in \mathcal{X}$ denote the continuous world state (object poses, robot
configuration, contact relations, etc.). The agent never observes $x$ directly;
at each time step $t$ it receives an observation $o_t \in \mathcal{O}$, which
may be an RGB image, a rendered simulator frame, or a natural-language description
of the scene.

Following the TAMP convention \cite{garrett2021integrated}, each continuous world
state $x_i$ admits a symbolic state representation
\[
  s_i=\sigma(x_i) \;=\; \{ f \in \mathcal{F} \mid f \text{ holds in } x_i \},
\]
where $\mathcal{F}$ is a domain-dependent set of grounded predicates
(e.g.\ \texttt{holding(none)}, \texttt{on(block1, table)}).
Because the true state $x$ is unobserved, we should deal with a symbolic belief state
$\hat{b}_t$ that represents the believe over which fluents currently hold:
\[
  \hat{b}_t \;=\; \bigl\{(f,\; p_t(f)) \;\big|\; f \in \mathcal{F}\bigr\},
  \;\;
\]
where $p_t(f)$ is the estimated probability that the fluent holds at time $t$. This belief is estimated by a VLM grounder
$\mathcal{M}_{\mathrm{VLM}}$, which maps the current observation to grounded
fluents.

In standard TAMP, actions are described by symbolic operators with explicit
preconditions and effects \cite{garrett2021integrated}. In our setting, no such
model is given a priori. Instead, we assume access to a set of closed-loop
goal-conditioned controllers $\mathcal{C} = \{ c_1, \ldots, c_K \}$, each
corresponding to a pretrained robotic skill (e.g.\ \texttt{pick},
\texttt{place}, \texttt{look-at}).

We define a Partially Observable TAMP problem as a tuple
$\mathcal{P} = (\mathcal{X},\, \mathcal{O},\, \hat{b}_0,\, \mathcal{C},\,
\mathcal{G})$,
where $\mathcal{X}$ is the continuous world-state space, $\mathcal{O}$ is the
observation space, $\hat{b}_0$ is the initial symbolic belief estimated from the
first observation $o_0$ via $\mathcal{M}_{\mathrm{VLM}}$, $\mathcal{C}$ is the
set of closed-loop controllers, and $\mathcal{G}$ is a goal condition expressed
as a conjunction of fluents that must hold with probability $\geq \delta$.

The agent must compute a plan
\[
  \pi \;=\; \bigl[(c_1, \theta_1),\; (c_2, \theta_2),\; \ldots,\;
                   (c_n, \theta_n)\bigr]
\]
such that the precondition of each $c_i$ is satisfied in $\hat{b}_{i-1}$ with
sufficiently high probability, all produced trajectories are geometrically
feasible, and executing $\pi$ from $\hat{b}_0$ yields a final belief satisfying
$\mathcal{G}$.

In this work, we instantiate this framework under deterministic grounding ($p_t(f) \in \{0,1\}$) and treat the simulator as the transition model, deferring belief-state maintenance and continuous feasibility to future work.

\section{Method}
\label{sec:method}
\begin{figure*}[thpb]
        \centering
        \includegraphics[width=\textwidth]{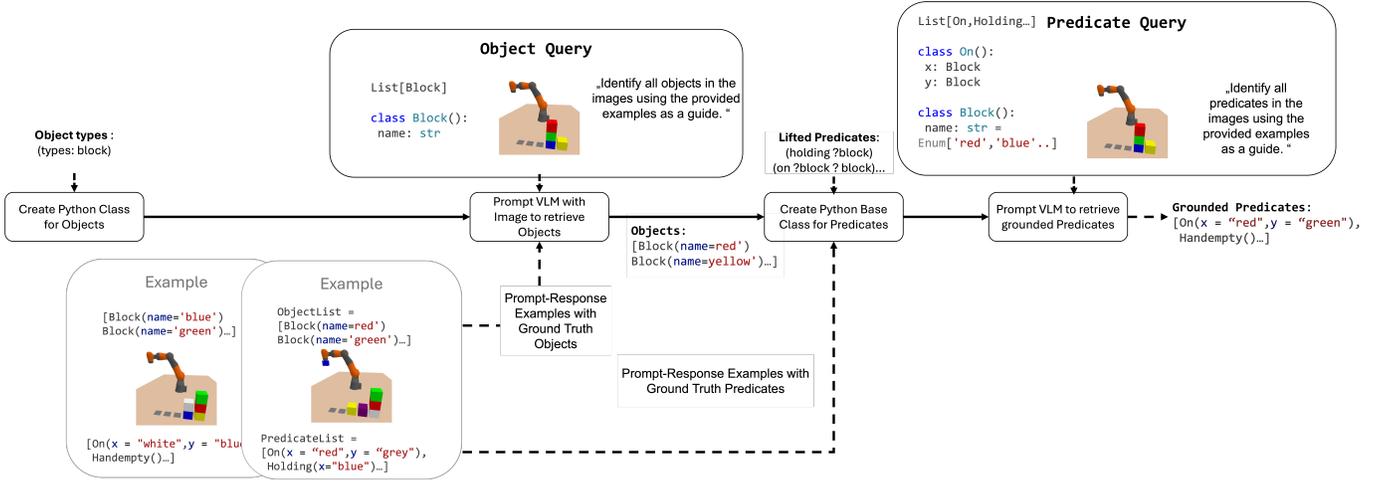}
        \caption{Flowchart of grounding from images or text to symbolic states.}
\label{fig:use_case_diagram}
\end{figure*}

We introduce SYMBOLIZER, a domain-agnostic framework composed of a grounding module and a planning module that enables model-free symbolic planning without an explicit action model. The grounding module constrains a vision-language model with a structured, first-order relational representation to ground vision or text inputs into symbolic states suitable for planning. A simulator is used as a black-box transition model that, given a state, generates successor states without an explicit specification of symbolic action model or their effects. The planner solves the resulting problem by systematic search in the induced state space, yielding a general, domain-independent problem-solving procedure. In this way, SYMBOLIZER turns raw observations into planning problems that can be addressed by a general solver, without domain-specific knowledge or manual modeling.

\subsection{SYMBOLIZER}
VLMs exhibit strong zero-shot perception capabilities, allowing them to interpret and describe previously unseen states without task-specific modeling. However, the naive descriptions
 are inherently flexible, with the same object often referred to in different ways (e.g., “red”, “red block”, or “red square”), and thus do not provide the precise and consistent symbolic representations required for planning over longer horizons. Existing approaches address this by constraining outputs to a fixed vocabulary or by querying individual predicates using domain knowledge, but this requires prior specification and limits zero-shot use. Moreover, such constraints do not guarantee well-formed representations and lead to representations that grow with the number of possible atoms, rendering domains with large state spaces, such as Blocksworld, infeasible beyond moderate sizes. To address this, we leverage the ability of VLMs to generate consistent structured outputs as Python objects from given class definitions. SYMBOLIZER automatically constructs these classes to encode a first-order relational representation of the domain. The resulting classes capture object types, their attributes, and the relations among objects. These classes constrain the VLM while preserving the expressivity of the state space, ensuring that generated states are well-formed with respect to object types and predicate arguments. This is achieved by requiring the model to produce outputs that conform to a typed object structure, where only valid fields and argument types are permitted. As a result, invalid symbols, inconsistent naming, and ill-typed predicates are ruled out at inference time. The model is provided with a generic, task-independent prompt and a small set of in-context examples, removing the need for task-specific prompt engineering. From these, the VLM infers the use of object types and predicates and generates consistent symbolic states.\\
\subsubsection{Grounding States}
SYMBOLIZER can ground symbolic states from either images or text. In both cases, objects are first extracted and then used to ground the final predicates. For the structured output to work effectively, the schema must cover all possible objects and predicates, without being too bloated, as unnecessary input can confuse the model. To strike this balance, we avoid creating a separate class for each type. Instead, we define a single Object class with a type field, selected out of a automatically generated list from the lifted representation, and a name field, which is a free-text entry inferred by the VLM from in-context examples. Object names are not fixed a priori and can, in principle, be chosen freely by the VLM. In practice, however, they are inferred from the model’s outputs and shaped by the naming patterns present in the in-context examples. While this does not impose explicit constraints at the level of object generation, the same names are subsequently used in predicates and goals, which enforces a consistent symbolic vocabulary across the representation.
To ground predicates, we constrain the space of valid predicates through the objects and their types. Concretely, after extracting the set of objects in the previous step, we group them by type and, for each type, provide the list of object names that belong to it. Predicate definitions then specify, for each argument, the type of object that is permitted in that position. At inference time, the VLM is restricted to not introduce new symbols or arbitrary combinations. Instead, for each predicate argument, it selects a name from the list associated with the required type. In this way, the space of possible predicates is implicitly defined by the set of object names available for each type, and the type signature of each predicate. This avoids the need to explicitly enumerate all possible atoms, while ensuring that all generated predicates are well-typed and refer only to previously grounded objects.\\
With these classes, arbitrarily complex predicates can be defined. To ensure semantic consistency without explicitly querying every possible predicate for truth \cite{zhangDKPROMPTDomainKnowledge2024}, we use in-context learning \cite{dong2024survey}. Each prompt contains up to ten examples, where each example pairs either an image or a natural-language description with a JSON-serialized Python object. To preserve domain-agnostic applicability, the system prompt is kept deliberately sparse and contains only a brief instruction to follow the provided examples and identify the relevant objects or predicates. The remainder of the prompt consists entirely of examples. This streamlined setup enables the model to ground initial, goal, or current states from either simulator-generated images or natural-language inputs
\subsubsection{Predicting Successor States}
With this approach, \gls{grounder_name} can also be used to predict successor states based on a given current state and action. The objects are derived from the input state, and the same object classes are reused to ensure consistency. The prompt remains minimal and consists solely of in-context examples. Each example is a triplet, with the  JSON-serialized Python object representing a state, an action, and the corresponding predicted next state. 
\subsubsection{Planning}
Our goal is to enable planning without requiring prior knowledge about the environment beyond the generated state representation. This allows the approach to operate even under partial observability or stochastic action outcomes, as no explicit action models are assumed. We adopt a heuristic-based search strategy where the primary heuristic is goal count, i.e., the number of unsatisfied goal predicates. However, in problem settings where the goal is not composed of multiple predicates, goal count alone provides limited guidance. In such cases, width is used as a tie-breaking criterion, favoring states that introduce novel combinations of predicates and thereby encouraging exploration. As a baseline, we also consider using a VLM as a heuristic.

\section{Experiments}

% =========================
% tables.tex
% =========================
\newcommand{\tableObjectGroundingCombined}{%
\begin{table}[!t]
\centering
\caption{Object grounding performance (F1). Using the same Gemini 3.1 Flash Lite, \textsc{Symbolizer} surpasses \textsc{ViLaIn} with two retries on every \textsc{ProDG} domain. Performance is near-perfect on custom domains. Best per row in bold.}
\label{tab:object_grounding_combined}
\setlength{\tabcolsep}{3.2pt}
\footnotesize
\resizebox{\columnwidth}{!}{%
\begin{tabular}{llcccccccc}
\toprule
 \multirow{2}{*}{Framework} & \multirow{2}{*}{Domain} & \multicolumn{2}{c}{ViLaIn (GPT-4)$^\dagger$} & \multicolumn{2}{c}{ViLaIn (Gem.~3.1-FL)} & \multicolumn{3}{c}{Symbolizer (Ours)}\\
 \cmidrule(lr){3-4} \cmidrule(lr){5-6} \cmidrule(lr){7-9}
 & & no retry & two retries & no retry & two retries & Gem.3.1-Pro & Gem.3.1-FL & Mistral Small\\
\midrule
ProDG          & Blocks      & 0.970 & 0.980 & 0.929 & 0.948 & 0.965 & \textbf{0.983} & \textbf{0.983} \\
ProDG          & Cooking     & 0.960 & 0.960 & 0.957 & 0.957 & \textbf{1.000} & \textbf{1.000} & 0.996 \\
ProDG          & Hanoi       & 0.890 & 0.890 & 0.897 & 0.900 & 0.943 & \textbf{0.999} & 0.927 \\
\midrule
PDDLGym        & Blocksworld & --- & ---   & 0.714 & 0.735 & \textbf{1.000} & 0.979 & \textbf{1.000} \\
PDDLGym        & Hanoi       & ---   & ---   & 0.215 & 0.684 & \textbf{1.000} & \textbf{1.000} & \textbf{1.000} \\
PDDLGym        & Hanoi Color & ---   & ---   & 1.000 & 1.000 & \textbf{1.000} & \textbf{1.000} & \textbf{1.000} \\
PyBullet       & Blocks      & ---   & ---   & 0.690 & 0.744 & \textbf{1.000} & 0.987 & 0.964 \\
PyBullet       & Hanoi       & ---   & ---   & 0.000 & 0.643 & \textbf{1.000} & 0.998 & 0.994 \\
Real Images    & Blocksworld & ---   & ---   & 0.474 & 0.560 & \textbf{1.000} & \textbf{1.000} & 0.958 \\
Kitchen-Worlds & ---         & ---   & ---   & --- & --- & \textbf{1.000} & 0.839 & \textbf{1.000} \\
\bottomrule
\multicolumn{9}{l}{\scriptsize $^\dagger$Reported.}
\end{tabular}%
}
\end{table}%
}

\newcommand{\tablePredicateGroundingCombined}{%
\begin{table}[!t]
\centering
\caption{Predicate grounding performance (F1). Even Mistral Small surpasses \textsc{ViLaIn} with two retries on all \textsc{ProDG} domains. Gemini 3.1 Pro leads on most custom domains. \textsc{Kitchen-Worlds} remains the most challenging setting. Best per row in bold.}
\label{tab:predicate_grounding_combined}
\setlength{\tabcolsep}{3.2pt}
\footnotesize
\resizebox{\columnwidth}{!}{%
\begin{tabular}{llcccccccc}
\toprule
 \multirow{2}{*}{Framework} & \multirow{2}{*}{Domain} & \multicolumn{2}{c}{ViLaIn (GPT-4)$^\dagger$} & \multicolumn{2}{c}{ViLaIn (Gem.~3.1-FL)} & \multicolumn{3}{c}{Symbolizer (Ours)}\\
 \cmidrule(lr){3-4} \cmidrule(lr){5-6} \cmidrule(lr){7-9}
 & & no retry & two retries & no retry & two retries & Gem.3.1-Pro & Gem.3.1-FL & Mistral Small\\
\midrule
ProDG          & Blocks      & 0.790 & 0.740 & 0.579 & 0.616 & 0.965 & \textbf{0.977} & 0.925 \\
ProDG          & Cooking     & 0.690 & 0.910 & 0.846 & 0.846 & \textbf{1.000} & \textbf{1.000} & 0.964 \\
ProDG          & Hanoi       & 0.470 & 0.520 & 0.658 & 0.641 & 0.886 & \textbf{0.973} & 0.784 \\
\midrule
PDDLGym        & Blocksworld & ---   & ---   & 0.357 & 0.376 & \textbf{1.000} & 0.981 & 0.969 \\
PDDLGym        & Hanoi       & ---   & ---   & 0.011 & 0.279 & \textbf{0.986} & 0.898 & 0.846 \\
PDDLGym        & Hanoi Color & ---   & ---   & 0.808 & 0.794 & 0.868 & 0.842 & \textbf{0.889} \\
PyBullet       & Blocks      & ---   & ---   & 0.331 & 0.368 & \textbf{1.000} & 0.992 & 0.881 \\
PyBullet       & Hanoi       & ---   & ---   & 0.130 & 0.296 & \textbf{0.950} & 0.763 & 0.841 \\
Real Images    & Blocksworld & ---   & ---   & 0.264 & 0.323 & \textbf{1.000} & \textbf{1.000} & 0.883 \\
Kitchen-Worlds & ---         & ---   & ---   & --- & --- & \textbf{0.851} & 0.695 & 0.754 \\
\bottomrule
\multicolumn{9}{l}{\scriptsize $^\dagger$Reported.}
\end{tabular}%
}
\end{table}%
}

\newcommand{\tableGoalGroundingCombined}{%
\begin{table}[!t]
\centering
\caption{Goal grounding performance (F1). Results are more variable, reflecting greater compositional reasoning demands. Performance is model-sensitive, with smaller models degrading on \textsc{Real Images}, indicating goal grounding as the current bottleneck. Best per row in bold.}
\label{tab:goal_grounding_combined}
\setlength{\tabcolsep}{3.2pt}
\footnotesize
\resizebox{\columnwidth}{!}{%
\begin{tabular}{llcccccccc}
\toprule
 \multirow{2}{*}{Framework} & \multirow{2}{*}{Domain} & \multicolumn{2}{c}{ViLaIn (GPT-4)$^\dagger$} & \multicolumn{2}{c}{ViLaIn (Gem.~3.1-FL)} & \multicolumn{3}{c}{Symbolizer (Ours)}\\
 \cmidrule(lr){3-4} \cmidrule(lr){5-6} \cmidrule(lr){7-9}
 & & no retry & two retries & no retry & two retries & Gem.3.1-Pro & Gem.3.1-FL & Mistral Small\\
\midrule
ProDG          & Blocks      & 0.930 & 0.900 & \textbf{0.958} & 0.943 & 0.944 & 0.944 & 0.934 \\
ProDG          & Cooking     & 0.880 & 0.930 & 0.923 & 0.923 & \textbf{1.000} & \textbf{1.000} & 0.962 \\
ProDG          & Hanoi       & 0.020 & 0.330 & 0.407 & 0.390 & \textbf{0.710} & 0.586 & 0.198 \\
\midrule
PDDLGym        & Blocksworld & ---   & ---   & \textbf{1.000} & 0.956 & \textbf{1.000} & 0.964 & 0.890 \\
PDDLGym        & Hanoi       & ---   & ---   & 0.412 & 0.655 & \textbf{1.000} & \textbf{1.000} & 0.682 \\
PDDLGym        & Hanoi Color & ---   & ---   & \textbf{1.000} &\textbf{ 1.000} & \textbf{1.000} & \textbf{1.000} & 0.857 \\
PyBullet       & Blocks      & ---   & ---   & \textbf{1.000} & 0.864 & \textbf{1.000} & \textbf{1.000} & 0.974 \\
PyBullet       & Hanoi       & ---   & ---   & \textbf{1.000} & 0.971 & \textbf{1.000} & 0.989 & 0.969 \\
Real Images    & Blocksworld & ---   & ---   & \textbf{1.000} & 0.802 & \textbf{1.000} & \textbf{1.000} & 0.787 \\
Kitchen-Worlds & ---         & ---   & ---   & --- & --- & \textbf{1.000} & \textbf{0.800} & 0.700 \\
\bottomrule
\multicolumn{9}{l}{\scriptsize $^\dagger$Reported.}
\end{tabular}%
}
\end{table}%
}

\newcommand{\tableProDGPlanning}{%
\begin{table}[!t]
\centering
\caption{Planning success rate (\%) with symbolic action model. Symbolizer outperforms ViLaIn in problem generation, especially for novel visualizations. Best in bold.}
\label{tab:prodg_planning}
\setlength{\tabcolsep}{2.5pt}
\footnotesize
\resizebox{\columnwidth}{!}{%
\begin{tabular}{llccccccc}
\toprule
\multirow{2}{*}{Framework} & \multirow{2}{*}{Domain} & \multicolumn{3}{c}{Symbolizer (Ours)} & \multicolumn{2}{c}{ViLaIn (Gem.~3.1-FL)} & \multicolumn{2}{c}{ViLaIn (GPT-4)$^\dagger$} \\
\cmidrule(lr){3-5} \cmidrule(lr){6-7} \cmidrule(lr){8-9}
 & & Gem.3.1-Pro & Gem.3.1-FL & Mistral Small & no retry & 2 retries & no retry & 2 retries \\
\midrule
ProDG   & Blocks  & \textbf{90.0} & \textbf{90.0} & 53.3 & 37.0 & 79.0 & 36.0 & 40.0 \\
ProDG   & Cooking & 96.7 & \textbf{100.0} & 66.7 & 70.0 & 90.0 & 9.0 & 71.0 \\
ProDG   & Hanoi   & \textbf{50.0} & 20.0 & 0.0 & 20.0 & 20.0 & 0.0 & 12.0 \\
\midrule
PDDLGym & Blocks  & \textbf{100.0} & 96.0 & 32.0 & 10.0 & 10.0 & --- & --- \\
PDDLGym & Hanoi   & \textbf{100.0} & 92.0 & 40.0 & 0.0 & 7.0 & --- & --- \\
PyBullet & Blocks & \textbf{100.0} & \textbf{100.0} & 36.0 & 0.0 & 0.8 & --- & --- \\
PyBullet & Hanoi  & \textbf{80.0} & 48.0 & 20.0 & 0.0 & 1.2 & --- & --- \\
\bottomrule
\multicolumn{9}{l}{\scriptsize $^\dagger$Reported.}
\end{tabular}%
}
\end{table}%
}

\newcommand{\tablePlanningOwnDomains}{%
\begin{table}[!t]
\centering
\caption{Planning success rate on \textsc{PDDLGym} and custom simulator domains. Grounded symbolic search outperforms direct VLM planning by a wide margin. Adding a VLM-based heuristic yields no meaningful improvement, indicating that accurate grounding paired with systematic search is the primary driver. Best per row in bold.}
\label{tab:planning_own_domains}
\setlength{\tabcolsep}{3pt}
\footnotesize
\resizebox{\columnwidth}{!}{%
\begin{tabular}{llccc}
\toprule
Framework & 
\begin{tabular}{@{}l@{}}Domain / \\ Task Family\end{tabular} & 
\begin{tabular}{@{}c@{}}Symbolizer$^\ast$ \end{tabular} & 
\begin{tabular}{@{}c@{}}Symbolizer$^\ast$ \\ + VLM Heuristic \end{tabular} & 
\begin{tabular}{@{}c@{}}Direct VLM \\ Plan \end{tabular} \\
\midrule
PDDLGym  & Blocksworld & 0.95 & \textbf{1.00} & 0.40 \\
PDDLGym  & Hanoi       & \textbf{1.00} & \textbf{1.00} & 0.40 \\
PDDLGym  & Hanoi Color & \textbf{1.00} & \textbf{1.00} & 0.12 \\
PyBullet & Blocks      & \textbf{1.00} & \textbf{1.00} & 0.44 \\
PyBullet & Hanoi       & \textbf{0.70} & \textbf{0.70} & 0.20 \\
Kitchen  & Kitchen     & \textbf{0.60} & \textbf{0.60} & 0.40 \\
\bottomrule
\end{tabular}%
}
\end{table}%
}

\newcommand{\tableViPlanMain}{%
\begin{table}[!t]
\centering
\caption{Planning success rate on \textsc{ViPlan}. \textsc{Symbolizer} achieves the best result on every row using only grounding and systematic search, compared against the strongest per-task configuration of \textsc{ViPlan} and \textsc{Dream-VL}. \textsc{Symbolizer} and our \textsc{ViPlan} reruns use Gemini 3.1 Flash Lite. Originally reported numbers in parentheses. Best per row in bold.}\label{tab:viplan_main}
\setlength{\tabcolsep}{3pt}
\footnotesize
\resizebox{\columnwidth}{!}{%
\begin{tabular}{llcccccc}
\toprule
Problem &
Setting &
\begin{tabular}{@{}c@{}}Symbolizer$^\ast$\end{tabular} &
\begin{tabular}{@{}c@{}}ViPlan-\\ Planner\end{tabular} &
\begin{tabular}{@{}c@{}}ViPlan-\\ Grounder\end{tabular} &
\begin{tabular}{@{}c@{}}Dream-VL \\ Grounding\end{tabular} &
\begin{tabular}{@{}c@{}}Dream-VL \\ Planning\end{tabular} \\
\midrule
Blocksworld & Simple & \textbf{1.00} & 0.96 (0.84) & 0.80 (\textbf{1.00}) & 0.16 & 0.00 \\
Blocksworld & Medium & \textbf{1.00} & 0.92 (0.48) & 0.20 (0.76) & 0.00 & 0.00 \\
Blocksworld & Hard   & \textbf{0.76} & 0.44 (0.12) & 0.00 (0.48) & 0.00 & 0.00 \\
Household   & Simple & \textbf{0.92} & 0.08 (0.92) & 0.44 (0.28) & 0.04 & 0.00 \\
Household   & Medium & \textbf{0.88} & 0.00 (0.52) & 0.48 (0.04) & 0.00 & 0.00 \\
Household   & Hard   & \textbf{0.40} & 0.00 (0.36) & 0.00 (0.08) & 0.04 & 0.00 \\
\bottomrule
\end{tabular}%
}
\end{table}%
}

We evaluate \textsc{Symbolizer} along two axes. First, we assess the quality of the symbolic grounding produced by the VLM, measuring how accurately it recovers objects, predicates, and goals from visual and textual observations. Second, we test whether the resulting representations are accurate enough to enable effective planning. We begin by describing the domains and benchmarks used across both evaluations, then present grounding quality and planning performance in turn.

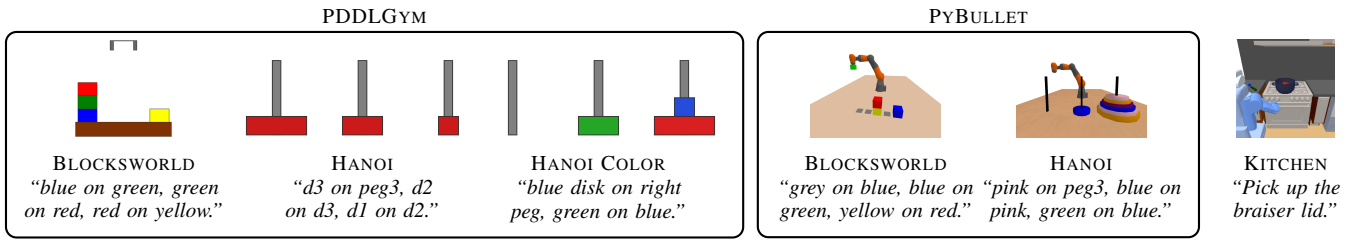
\begin{figure*}[t]
    \centering
    \footnotesize
    \begin{tikzpicture}[
        font=\footnotesize,
        panel/.style={inner sep=0pt, outer sep=0pt, anchor=north west},
        caption/.style={align=center, text width=0.15\textwidth, anchor=north},
        groupbox/.style={draw, rounded corners, thick, inner sep=3pt}
    ]

    \def\imgh{13mm}
    \def\xgap{3mm}
    \def\ygap{1.2cm}

    \node[panel] (bw) at (0,0)
        {\includegraphics[height=\imgh]{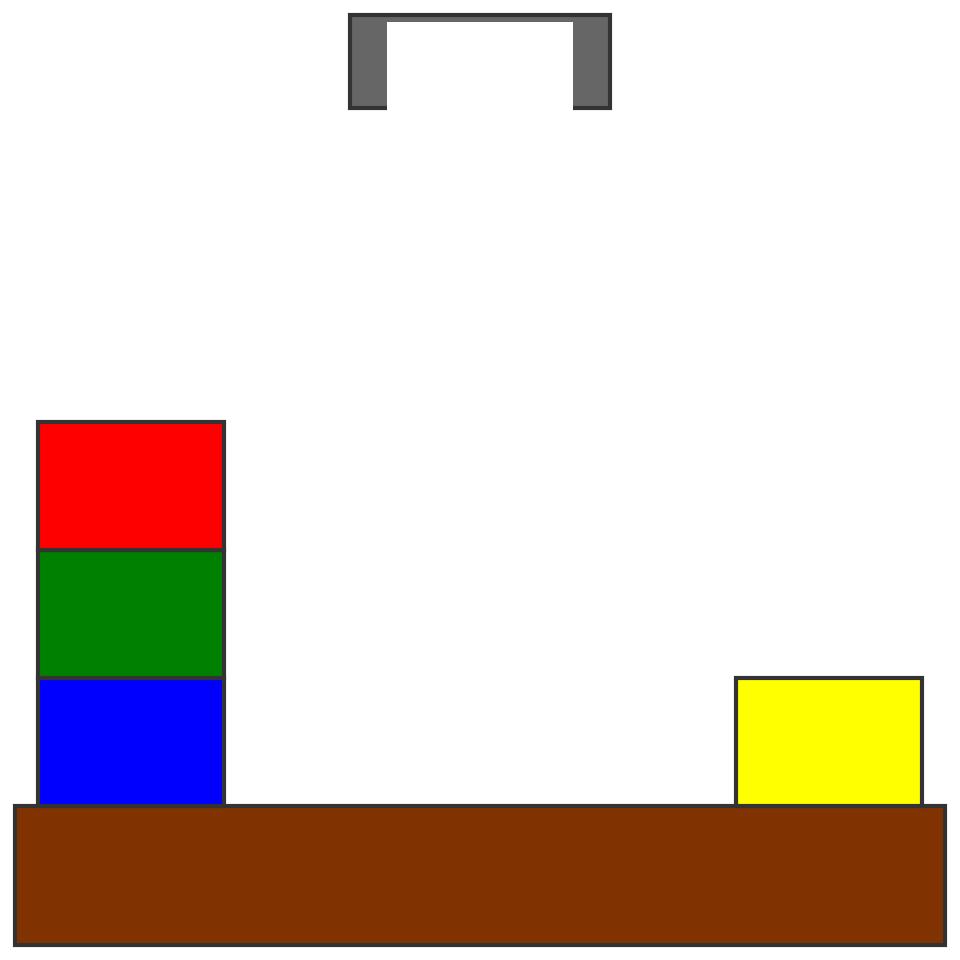}};
    \node[panel, right=7mm of bw] (hanoi)
        {\includegraphics[height=\imgh]{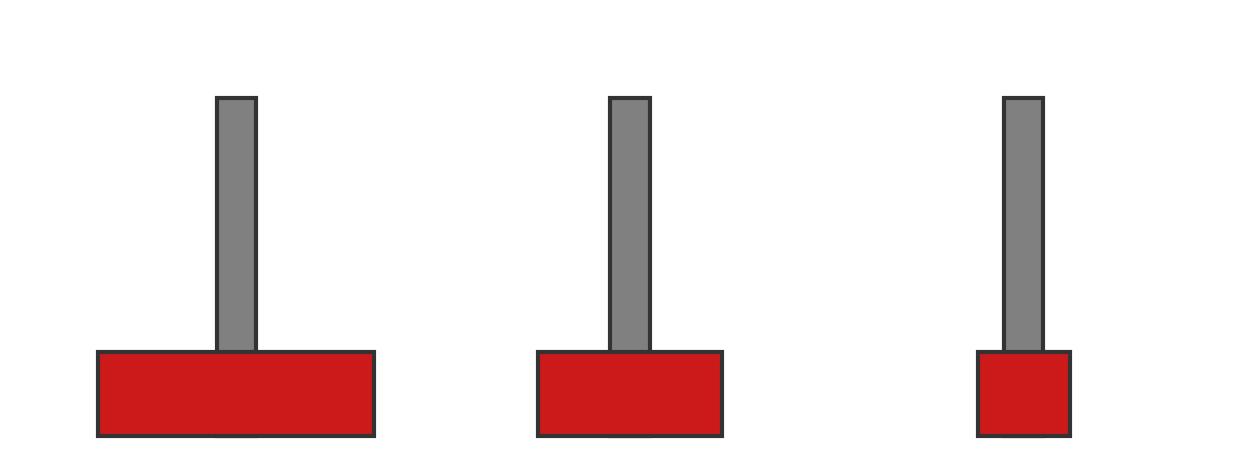}};
    \node[panel, right=-5mm of hanoi] (hcolor)
        {\includegraphics[height=\imgh]{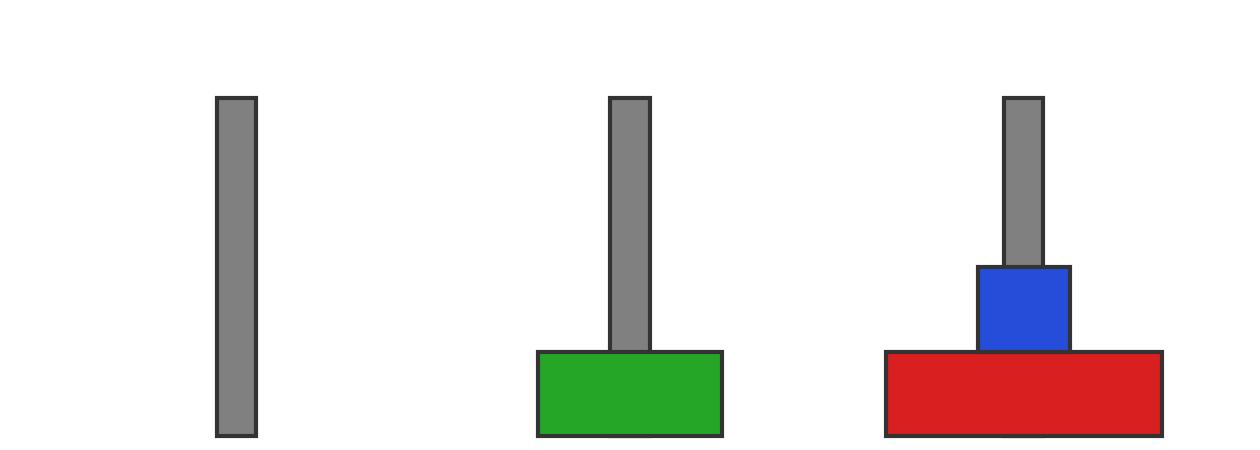}};
    \node[panel, right=10mm of hcolor] (pbw)
        {\includegraphics[height=\imgh]{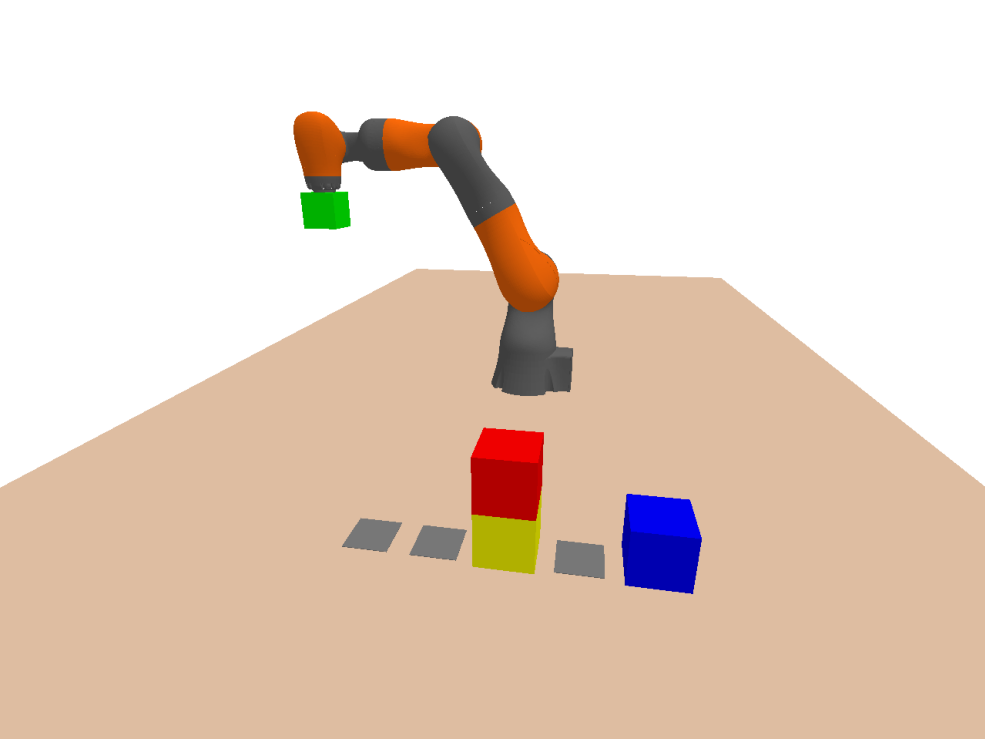}};
    \node[panel, right=10mm of pbw] (phanoi)
        {\includegraphics[height=\imgh]{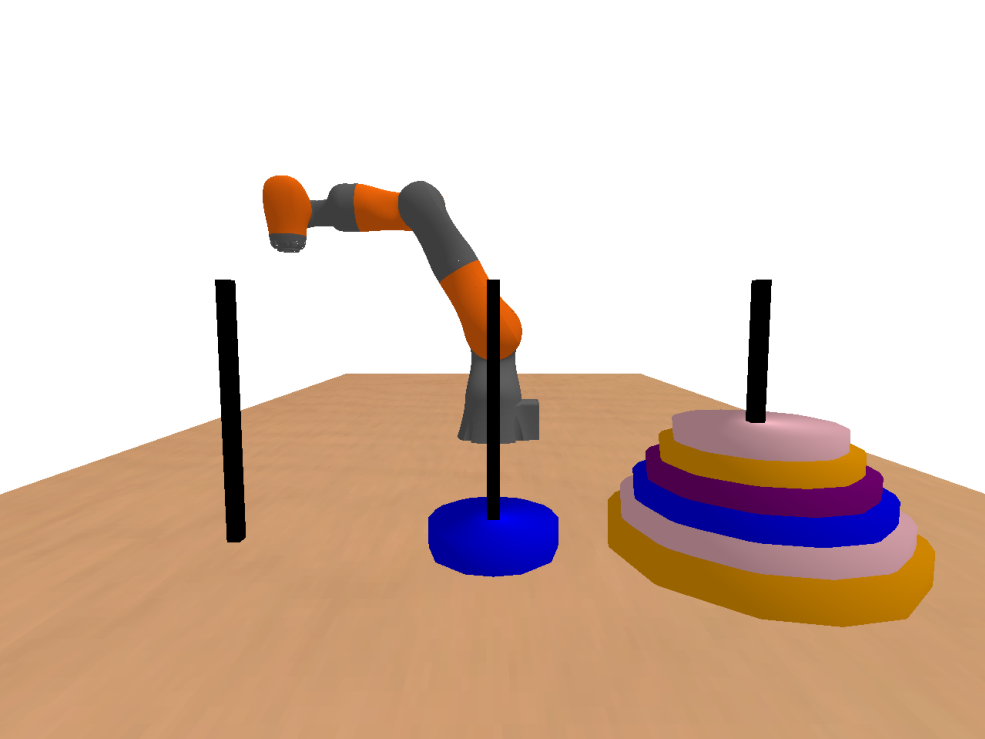}};
    \node[panel, right=12mm of phanoi] (kitchen)
        {\includegraphics[height=\imgh]{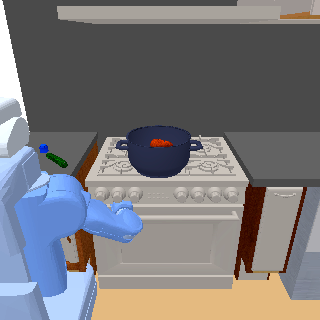}};

    \node[caption, below=4pt of bw.south] (cbw)
        {\textsc{Blocksworld}\\
        \textit{``blue on green, green on red, red on yellow.''}};

    \node[caption, below=4pt of hanoi.south] (chanoi)
        {\textsc{Hanoi}\\
        \textit{``d3 on peg3, d2 on d3, d1 on d2.''}};

    \node[caption, below=4pt of hcolor.south] (chcolor)
        {\textsc{Hanoi~Color}\\
        \textit{``blue disk on right peg, green on blue.''}};

    \node[caption, below=4pt of pbw.south] (cpbw)
        {\textsc{Blocksworld}\\
        \textit{``grey on blue, blue on green, yellow on red.''}};

    \node[caption, below=4pt of phanoi.south] (cphanoi)
        {\textsc{Hanoi}\\
        \textit{``pink on peg3, blue on pink, green on blue.''}};

    \node[caption, below=4pt of kitchen.south] (ckitchen)
        {\textsc{Kitchen}\\
        \textit{``Pick up the braiser lid.''}};

    \node[groupbox, fit=(bw)(cbw)(hanoi)(chanoi)(hcolor)(chcolor),
          label={north:\textsc{PDDLGym}}] {};

    \node[groupbox, fit=(pbw)(cpbw)(phanoi)(cphanoi),
          label={north:\textsc{PyBullet}}] {};

    \end{tikzpicture}
    
    \caption{Example observations and goal specifications from our custom evaluation domains. (a)--(c): \textsc{PDDLGym} (2D sprites). (d)--(f): \textsc{PyBullet} (3D rendering). Goals shown in italics.}
    \label{fig:own_domains}
\end{figure*}

Our evaluation spans both custom domains and established external benchmarks. The custom domains include \textsc{Blocksworld} and \textsc{Hanoi}, each implemented in two synthetic simulators, \textsc{PDDLGym} with 2D sprites and a custom \textsc{PyBullet} environment with 3D physics rendering. For \textsc{PDDLGym}, we additionally include a \textsc{Hanoi Color} variant where disks are distinguished by color rather than size. We further include \textsc{Kitchen-Worlds}, which introduces a broader set of objects and predicates. Ground-truth symbolic states are available from each simulator, allowing evaluation at each stage of the grounding pipeline. Figure~\ref{fig:own_domains} shows example observations and goal specifications from these domains. For external benchmarks, we use \textsc{ProDG} for grounding evaluation and \textsc{ViPlan} for planning evaluation. \textsc{ProDG}, introduced by the \textsc{ViLaIn} authors, provides visually grounded \textsc{Cooking} and \textsc{Hanoi} problems with ground-truth annotations for objects, predicates, and goals. \textsc{ViPlan} evaluates end-to-end visual planning in a photorealistic \textsc{Blocksworld} simulator and in \textsc{Household} tasks with partial observability, each at three difficulty levels. Notably, the \textsc{Household} domain provides initial and goal states as natural language descriptions rather than images, which lets us test whether our grounding pipeline generalizes beyond visual input.

\subsection{Grounding Evaluation}

We evaluate grounding on all custom domains, that is \textsc{Blocksworld} and \textsc{Hanoi} in both \textsc{PDDLGym} and \textsc{PyBullet}, \textsc{Hanoi Color}, and \textsc{Kitchen-Worlds}, as well as on the external \textsc{ProDG}  benchmark, published together with the ViLaIn paper \cite{shirai2024vision}. Together, these span a range of perceptual complexity, from simple 2D sprites to 3D rendered scenes to real images. We measure F1 scores for object, predicate, and goal extraction across three VLMs of varying capacity, namely Gemini 3.1 Pro, Gemini 3.1 Flash Lite and Mistral Small 2503, included to provide an open-weights baseline for reproducibility. All three models receive the same domain-agnostic prompt with three in-context examples and no domain-specific tuning. Following our grounding pipeline, objects are extracted first, and the resulting object set is then provided as input to the predicate grounding stage, which generates predicates sequentially conditioned on the available objects. Goals are grounded in a separate step. F1 is computed per instance by comparing predicted and ground-truth sets element-wise, treating each object, predicate instance, or goal literal as an individual element, and then averaged across all instances.

On the \textsc{ProDG} dataset, we additionally compare against \textsc{ViLaIn} \cite{shirai2024vision} under two configurations: one without retries, approximating our single-pass setting where the VLM produces grounding in a single forward call, and one with two planner-feedback retries, corresponding to ViLaIn’s full multi-stage pipeline in which grounding errors identified by the planner are fed back for correction. On \textsc{ProDG}, we include the originally reported \textsc{ViLaIn} results using GPT-4 as well as our own rerun with Gemini 3.1 Flash Lite for a controlled same-model comparison. We further run \textsc{ViLaIn} with Gemini 3.1 Flash Lite on our custom domains. Beyond component-level metrics, we further use \textsc{ProDG} to evaluate whether each model can generate complete, well-formed PDDL problem files that can be consumed and solved by a classical planner, thereby assessing whether individually grounded components compose into a coherent symbolic representation rather than only measuring their accuracy in isolation.
\subsubsection{Object grounding}
Table~\ref{tab:object_grounding_combined} shows that object grounding is highly reliable across all models. On \textsc{ProDG}, all three models match or exceed \textsc{ViLaIn}, the current SOTA baseline, even when \textsc{ViLaIn} is given two planner-feedback retries. Using the same Gemini 3.1 Flash Lite backbone, \textsc{Symbolizer} surpasses \textsc{ViLaIn} on every \textsc{ProDG} domain, even when \textsc{ViLaIn} is given two planner-feedback retries. On our custom domains, performance is near-perfect across both simulators and real images, while \textsc{ViLaIn} with the same Gemini 3.1 Flash Lite shows substantially lower performance on \textsc{PDDLGym Blocksworld}, \textsc{Hanoi}, and \textsc{Real Images}. Since object grounding is a visual task, this gap reflects the cascading errors in \textsc{ViLaIn}’s multi-stage visual pipeline, where an open-vocabulary detector, a captioning model, and rule-based formatting each introduce failure points that a single structured VLM call avoids. Generalization performance varies across domains. The contrast between \textsc{Hanoi} and \textsc{Hanoi Color} is particularly revealing. When symbolic identifiers align with visually salient features such as color, all models achieve near-perfect F1. In contrast, when identifiers depend on size-based reasoning, performance becomes more variable and model-dependent. While larger models tend to perform better, scale alone does not explain the results, as the structured output schema constrains the output space enough for even smaller models to produce accurate object lists in most settings.

\tableObjectGroundingCombined

\subsubsection{Predicate grounding}
Predicate grounding is substantially harder than object grounding, as the model must recover relational structure rather than enumerate entities. Moreover, since our pipeline grounds predicates conditioned on the previously extracted objects, errors from the first stage propagate into this step. Table~\ref{tab:predicate_grounding_combined} reflects this through lower F1 scores and greater variance across models and domains. Nevertheless, on \textsc{ProDG}, all three models surpass \textsc{ViLaIn} with two retries on every domain. This includes Mistral Small, whose training data cuts off before the \textsc{ProDG} benchmark was published, ruling out memorization of the evaluation data. This demonstrates that our grounding scheme captures generalizable structure rather than relying on domain-specific memorization from the training data. On our custom domains, \textsc{ViLaIn} performs poorly across all settings, whereas all \textsc{Symbolizer} models maintain consistently strong performance. As with object grounding, predicate recovery requires visual understanding, the same cascading errors in \textsc{ViLaIn}'s visual pipeline that degrade object detection propagate further into predicate extraction, where the relational structure makes the task even less forgiving. Across custom domains, model scale plays a more visible role for predicate grounding than for object grounding. Gemini 3.1 Pro leads in most settings, consistent with the expectation that relational reasoning is more demanding and benefits from greater model capacity. The \textsc{Hanoi} domains are challenging across both simulators, largely due to static predicates that must be inferred but never change across states, making them difficult to acquire from few in-context examples. \textsc{Kitchen-Worlds} is the hardest setting overall. Objects are physically smaller and harder to distinguish visually, and the same static predicate problem occurs. On \textsc{Real Images Blocksworld}, all models achieve strong F1, confirming that predicate recovery transfers to non-synthetic scenes.
\tablePredicateGroundingCombined

\subsubsection{Goal grounding}
\tableGoalGroundingCombined
Goal grounding shows the largest variance across models and domains (Table~\ref{tab:goal_grounding_combined}). Unlike object and predicate grounding, goal grounding is text-only across all evaluated domains, requiring the model to infer target configurations from fewer examples without visual context. This favors larger models, with Gemini 3.1 Pro achieving near-perfect F1 while Mistral Small degrades substantially. Since \textsc{ViLaIn}'s goal generation also bypasses its visual pipeline entirely, both approaches are more comparable here, and the gap is smaller than for visual grounding. On \textsc{ProDG}, \textsc{Symbolizer} matches or outperforms \textsc{ViLaIn} across all domains. Goal errors compound from the object stage, and some static predicates such as \texttt{clear} are annotated inconsistently across \textsc{ProDG} instances. \textsc{ViLaIn}'s performance remains inconsistent with retries, and both in the reported GPT-4 results and our Gemini 3.1 Flash Lite rerun, adding retries can degrade goal quality, as corrective reprompting does not re-observe the scene and optimizes for planability rather than grounding accuracy. On more abstract domains such as \textsc{Hanoi}, \textsc{Symbolizer} maintains a clear advantage. Goal grounding is the current bottleneck of the pipeline, but since goals are grounded only once per problem instance, using a larger model is practical and incurs negligible additional cost.

\begin{figure}[t]
    \captionsetup[subfloat]{captionskip=2pt,farskip=2pt,nearskip=1pt}
    \centering
    \subfloat[\textsc{Blocksworld}\\\footnotesize\textit{``Stack: yellow over pink over green over red over purple over blue.''}\label{fig:domain_vilain_bw}]{%
        \includegraphics[width=0.32\columnwidth]{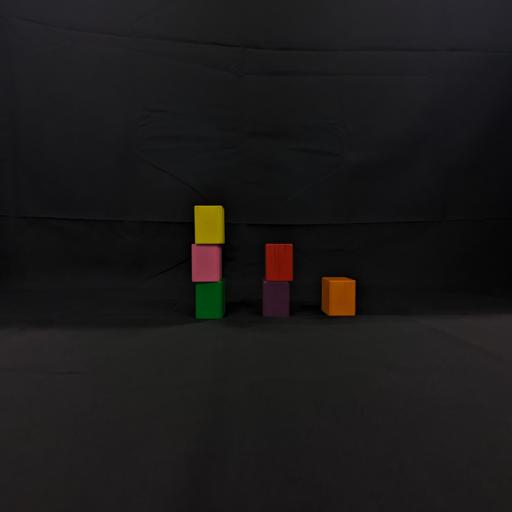}%
    }\hfill
    \subfloat[\textsc{Cooking}\\\footnotesize\textit{``Slice the cucumber and tomato, place in bowl.''}\label{fig:domain_vilain_cooking}]{%
        \includegraphics[width=0.32\columnwidth]{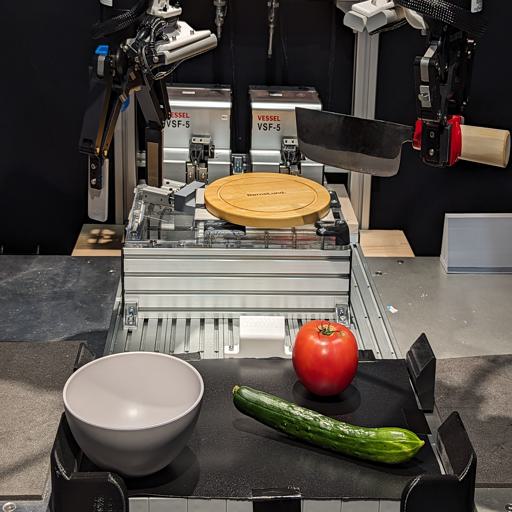}%
    }\hfill
    \subfloat[\textsc{Hanoi}\\\footnotesize\textit{``Move all disks to the rightmost peg.''}\label{fig:domain_vilain_hanoi}]{%
        \includegraphics[width=0.32\columnwidth]{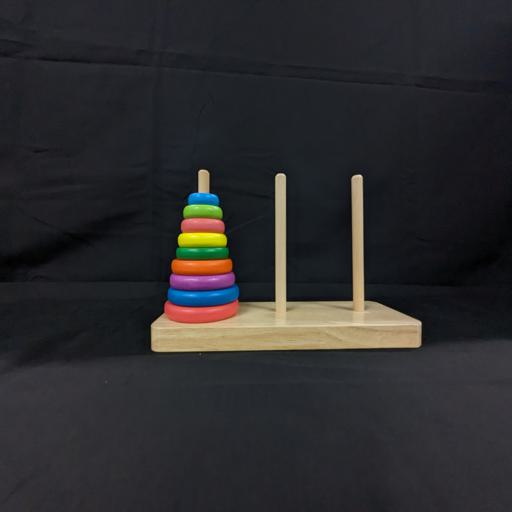}%
    }
    \caption{Example observations and goals from the \textsc{ProDG} benchmark (real images). These domains test grounding on physical scenes with diverse objects and relations.}
    \label{fig:prodg_domains}
\end{figure}

\subsubsection{Planning with Symbolic Action Model}
\tableProDGPlanning
On \textsc{ProDG}, both Gemini models outperform \textsc{ViLaIn} on all three domains, even against \textsc{ViLaIn} with two retries (Table~\ref{tab:prodg_planning}). This is notable because \textsc{ViLaIn} was specifically designed for this setting, relying on domain-specific object detection, captioning, and the full action model, while \textsc{Symbolizer} uses none of these. On our custom domains with the same model, the gap widens dramatically. \textsc{ViLaIn}'s pipeline does not generalize beyond its original benchmark, even within the same planning domains implemented in different simulators. \textsc{Symbolizer} generalizes across simulators without any domain knowledge or handcrafting, demonstrating that structured grounding transfers where pipeline-based approaches fail.

\subsection{Planning Evaluation}

We evaluate planning on all custom simulator domains, that is \textsc{Blocksworld} and \textsc{Hanoi} in both \textsc{PDDLGym} and \textsc{PyBullet}, \textsc{Hanoi Color}, and \textsc{Kitchen-Worlds}, as well as on the external \textsc{ViPlan} \cite{merler2026viplanbenchmarkvisualplanning} benchmark. Planning performance is reported as success rate, defined as the fraction of problems for which a valid plan reaching the goal is found. All \textsc{Symbolizer} planning experiments use Gemini 3.1 Flash Lite for grounding, as it offered the best trade-off between accuracy and inference cost among the models we tested. We use goal-count as the search heuristic, with width-based search \cite{lipovetzky2012width} as a tie-breaker in domains where goal-count alone produces many equally scored states.\\
The \textsc{ProDG} evaluation protocol tests the ability to ground initial states and goals into PDDL problem files, which are then paired with the GT domain file and solved with a classical planner. \textsc{ViLaIn} uses Fast Downward and we use A$^\ast$ with goal-count heuristic. We compare against the originally reported \textsc{ViLaIn} results using GPT-4 as well as our own rerun using Gemini 3.1 Flash Lite. We further apply the same protocol on our custom \textsc{PDDLGym} and \textsc{PyBullet} implementations of \textsc{Blocksworld} and \textsc{Hanoi}, comparing \textsc{Symbolizer} and \textsc{ViLaIn} both with Gemini 3.1 Flash Lite.\\
On our custom domains, we compare \textsc{Symbolizer} against two baselines that help isolate the contribution of symbolic structure. The first is VLM-guided search, which replaces the classical heuristic with model-based guidance within the same search framework, similar in spirit to \cite{brohan2023can,hazra2024saycanpay,lin2023text2motion}. The second is direct VLM planning, where the model receives the initial state image, the goal specification, and the domain description, and produces a plan in a single forward pass. Together, these baselines separate the VLM's perceptual capability, which is all \textsc{Symbolizer} relies on, from its ability to perform multi-step reasoning about domain dynamics.\\
On the \textsc{ViPlan} benchmark, we evaluate \textsc{Symbolizer} in a setting with photorealistic rendering and against established baselines from the literature. The benchmark was introduced alongside the \textsc{ViPlan} planner and grounder, which evaluated multiple VLMs including frontier models such as GPT 5.2 for both grounding and planning. Since different models perform best on different domains and difficulty levels, we compare \textsc{Symbolizer} against the best result across all their models for each task-difficulty combination, providing an upper bound on their approach and ensuring the strongest possible baseline. For a controlled comparison, we additionally ran the \textsc{ViPlan} framework with Gemini 3.1 Flash Lite, the exact model used in our \textsc{Symbolizer} experiments. We further compare against \textsc{Dream-VL}, which combines VLM-based grounding with a learned world model for planning.
\subsubsection{Planning on Our Domains}
\tablePlanningOwnDomains
Table~\ref{tab:planning_own_domains} isolates the contribution of symbolic structure by comparing grounded symbolic search against direct VLM planning and VLM-guided search within the same framework. Grounded symbolic search outperforms direct VLM planning by a wide margin across all domains. Adding a VLM-based heuristic yields no meaningful improvement, with results identical in five of six domains. The value lies in grounding and systematic search, not in model-guided state evaluation. The failure modes of the two approaches differ fundamentally. Direct VLM planning requires the model to predict action outcomes correctly at every step, a task that is inherently harder than perception and leaves far less room for error, since a single mistake invalidates the remainder of the plan. Symbolic search decouples perception from reasoning. Grounding errors affect the initial state representation, but search explores the state space systematically and can recover as long as the representation remains tractable and goal-relevant structure is preserved. For robotic task planning, this separation is practical. Perception is a well-studied sensor problem, while long-horizon combinatorial reasoning is better handled by established search algorithms with completeness guarantees.
\subsubsection{Planning on ViPlan}

\begin{figure}[t]
    \captionsetup[subfloat]{captionskip=2pt,farskip=2pt,nearskip=1pt}
    \centering
    \subfloat[\textsc{Blocksworld}\\\footnotesize\textit{``green on purple in c3, red on orange in c2.''}\label{fig:domain_viplan_bw}]{%
        \includegraphics[height=3cm]{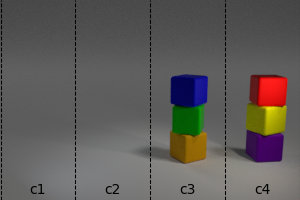}%
    }\hfill
    \subfloat[\textsc{Household}\\\footnotesize\textit{``close window\_1, window\_2, window\_3.''}\label{fig:domain_viplan_household}]{%
        \includegraphics[height=3cm]{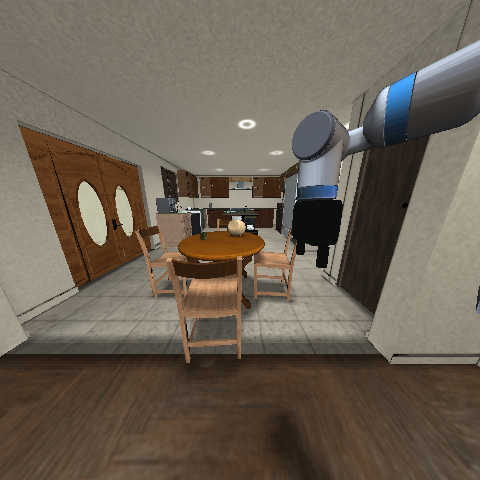}%
    }
    \caption{Example observations and goals from the \textsc{ViPlan} benchmark. \textsc{Blocksworld} uses photorealistic rendering, \textsc{Household} provides goals as natural language.}
    \label{fig:viplan_domains}
\end{figure}

\tableViPlanMain
Table~\ref{tab:viplan_main} shows that \textsc{Symbolizer} achieves the best result on every task-difficulty combination, outperforming all models and all methods using only grounding and systematic search. Gemini's knowledge cutoff predates the \textsc{ViPlan} code and paper, so no data leakage is possible. The reported \textsc{ViPlan} results represent the best score across over 20 models including frontier models such as GPT 5.2, while our rerun uses only Gemini 3.1 Flash Lite, a cost-focused model, thus achieving lower scores. Yet \textsc{Symbolizer} with the same cost-focused model outperforms even the best reported configuration on every setting. On \textsc{Blocksworld}, the advantage grows with problem difficulty. Systematic search benefits most where combinatorial depth is highest. \textsc{Household} is inherently harder, as the domain is broader, predicates are harder to learn from few examples, and objects and their states are visually harder to recognize. \textsc{Symbolizer} still leads at all difficulty levels, though hard \textsc{Household} remains challenging for all methods. Unlike all prior methods on this benchmark, our approach produces well-formed representations by construction. The only source of failure is perceptual misgrounding, not parsing or formatting errors. This is what makes search over large state spaces feasible in the first place.

\section{CONCLUSION}
We present SYMBOLIZER, a domain-agnostic framework for classical planning from images or text without a symbolic action model. By constraining VLM outputs to typed relational schemas, it generates well-formed symbolic states directly, eliminating post-processing, domain-specific prompts, and retries. Combined with a black-box simulator and systematic search, it enables planning with completeness guarantees beyond direct VLM approaches. Experiments show that even small open-source VLMs outperform prior multi-stage pipelines in both component metrics and end-to-end success, achieving SOTA results on \textsc{ViPlan}. The key insight is that modern VLMs already excel at grounding, while the main bottleneck is the lack of systematic search, and separating grounding from reasoning allows each to operate at its strength.

Several directions remain open. First, treating the VLM as a noisy sensor model and maintaining belief states over possible groundings could improve robustness under perceptual uncertainty. Second, closing the loop between grounding and domain generation, using VLMs to infer lifted action schemas from few interaction traces, would remove the remaining dependence on a pre-specified lifted representation. Third, applying SYMBOLIZER to real-world robotic domains, where sim-to-real transfer and partial observability introduce additional challenges, is a natural next step. Finally, the framework's generality suggests applicability beyond robotics to any domain expressible as a symbolic planning problem, such as logistics or scheduling, where visual or textual state descriptions are available but formal models are not.

\bibliographystyle{IEEEtran}
\bibliography{IEEEabrv,resources/references.bib}

\vfill

\end{document}